%% file: NoisePaper.tex
\documentclass[10pt,twocolumn,letterpaper]{article}

\usepackage{cvpr}
\usepackage{times}
\usepackage{epsfig}
\usepackage{graphicx}
\usepackage{amsmath}
\usepackage{amssymb}
\usepackage{subfig}

\newcommand{\hnm}{$H_0$}
\newcommand{\ham}{$H_1$}
\newcommand{\hn}{H_0}
\newcommand{\ha}{H_1}
\renewcommand{\vec}[1]{\mathbf{#1}}
\newcommand{\vym}{$\vec{y}$}

\newcommand{\vwm}{$\vec{w}$}
\newcommand{\vytm}{$\vec{y}_{train}$}

\newcommand{\vy}{\vec{y}}
\newcommand{\bk}{\vec{k}}

\newcommand{\vw}{\vec{w}}
\newcommand{\vyt}{\vec{y}_{train}}

\newcommand{\ip}[2]{<#1,#2>}
\newcommand{\sig}{\sigma^2_{\vw}}
\newcommand{\mv}{m_{\vw}}

\newcommand{\hypTest}{\overset{\hat{H}(\vy)=\ha}{\underset{\hat{H}(\vy)=\hn}{\gtreqless}}}
\newcommand{\revhypTest}{\overset{\hat{H}(\vy)=\hn}{\underset{\hat{H}(\vy)=\ha}{\gtreqless}}}

\cvprfinalcopy 

\def\cvprPaperID{0} 

\begin{document}

\title{A method of limiting performance loss of CNNs in noisy environments}

\author{James R. Geraci \\
Samsung Electronics Co,Ltd.\\
Seoul, South Korea\\
{\tt\small james.geraci@samsung.com} \and
\and
Parichay Kapoor\\
Samsung Electronics Co,Ltd.\\
Seoul, South Korea\\
{\tt\small pk.kapoor@samsung.com}
}

\maketitle

\input{abstract}

\input{intro}

\input{results}
\input{conclusions}

{\small
\bibliographystyle{ieee}
\bibliography{NoisePaper}
}

\end{document}

%% file: abstract.tex
\begin{abstract}
Convolutional Neural Network (CNN) recognition rates drop in the
presence of noise. We demonstrate a novel method of counteracting this
drop in recognition rate by adjusting the biases of the neurons in the
convolutional layers according to the noise conditions encountered at
runtime.

We compare our technique to training one network for all possible
noise levels, dehazing via preprocessing a signal with a denoising
autoencoder, and training a network specifically for each noise
level. Our system compares favorably in terms of robustness,
computational complexity and recognition rate.
\end{abstract}

%% file: intro.tex
\section{Background}
\label{sec:background}
Extracting or detecting signals in noise is a topic of long standing
interest in fields such as speech recognition
\cite{Cerisara200425,Gong:1995:SRN:205262.205266} and image processing
\cite{Besag86onthe}.  In the field of neural networks, denoising
autoencoders have been developed and their use to remove noise from
images has been suggested \cite{Vincent:2010:SDA:1756006.1953039}. In
the area of convolutional neural networks, operating in noisy
environments reduces recognition rates \cite{pinto_why_2008}. However,
training with noisy input improves the ability of the network to
detect signals in the presence of noise and such noisy trained
networks also perform well in the absence of noise
\cite{Sietsma:1991:CAN:104793.104807, Bishop:1995:TNE:211171.211185}.

How a CNN performs in the presence of noisy inputs is becoming
increasingly important due to their use in autonomous vehicles. In
that case, rain, snow, and fog make it difficult to detect and
identify other objects on the road. Snow also makes it difficult to
determine the vehicle's position relative to the road
\cite{FORDSNOW,MITLLGPR}.

The most obvious method is to train, perhaps smaller, networks for
each possible expected level of occlusion. In our experiments, this
method gives the best performance in terms of detection rate; however,
it is cumbersome in realize in practice because each noise level has
to have its own set of parameters. Loading and unloading all the
parameters for a large network at runtime could start to take
significant time and energy. Furthermore, storage space would have to
be provided for each set of parameters. For multiple copies of large
networks, this can easily require multi-GBs, so cost of storage memory
could become an issue.

To avoid having to deal with multiple sets of network parameters, one
common way of dealing with snow, rain and other forms of occlusion is
to train a network with a wide range, or mix, of representative input
conditions \cite{DBLP:BojarskiTDFFGJM16}. However, while this method
performs well over a wide range of noise levels, it does not perform
as well as other methods at the extreme edges of no noise and lots of
noise.

Our method requires the parameter sets at each extreme
of an expected occlusion/noise range. It receives a measurement of the
amount of occlusion in the environment and uses that measurement to
select which parameter set to use, the zero noise parameter set or the
max noise parameter set. It loads the parameters of the required
parameter set into the network, then adjusts the biases of the input
convolutional layer to help counteract the effects of noise.

To the best of our knowledge, there are no other works that
dynamically tune an individual part of a network at runtime to improve
performance in the presence of noise. In fact, while dehazing an image
seems to be an active area of research \cite{7407890}, there seems to
be little work with CNN based vision systems and noise. This could be
due to the lack of a standard noisy version of the KITTI
\cite{Geiger2013IJRR} dataset or similar benchmark dataset that would
make comparison across works easy.

For the works that do exist, some works fuse mutliple types of input
information to improve robustness to noise, \cite{1238422}. Other
works, such as \cite{7761510}, propose more sophisticated network
architectures. For the most part, previous works appear to consider
the network as a monolithic whole. For example,
\cite{DBLP:BojarskiTDFFGJM16} trains the whole network against a mixed
dataset. Our system is not mutually exclusive to these techniques and
can be used in conjunction with them.

In the following sections we describe a basic convolutional neural
network, review binary hypothesis testing, describe our method in more
detail and our experiments, their results, and present our
conclusions.

\subsection{Basic CNN Model}
\begin{figure}[t]
\begin{center}
\fbox{\rule{0pt}{2in}
   \includegraphics[width=0.8\linewidth]{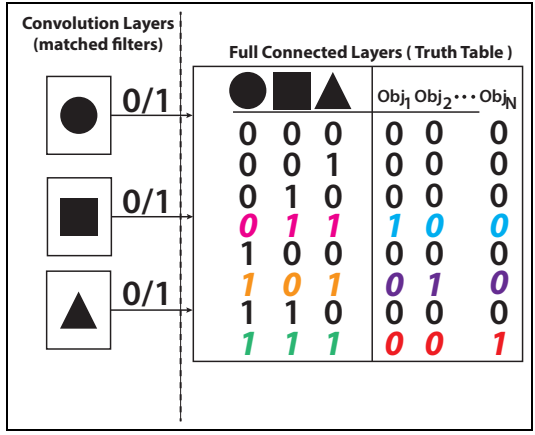}}
\end{center}
   \caption{A CNN can be thought of as a collection of matched filters
     attached to a truth table. In this example, presenting an image
     to the CNN projects that image onto the circle, square, triangle
     basis.}
\label{fig:BasicCNNmulti}
\end{figure}

A simple way to think about a CNN is as a collection of matched
filters (equivalently correlators) connected to a truth table,
Fig.\ref{fig:BasicCNNmulti}. Each correlator performs a binary
hypothesis test on each region of the input image to which it is
exposed. If, according to the binary hypothesis test, there is a match
between the object encoded in the filter and the area of the image the
filter is looking at, a `1' is output otherwise a `0' is output. If a
certain subset of the filters find their target shape, then a
particular line in a truth table is looked up. If there is an object
that corresponds to that line, then the output corresponding to that
object goes high. Any object that is not associated with some
combination of the objects in the matched filters cannot be detected.

\subsection{Binary Hypothesis Test}
\label{sec:BHT}
Eq.\ref{eq:hypotheses} shows the two hypotheses for our system. Here,
\hnm \ is the hypothesis that the observed data, \vym, is just random
noise, \vwm. \ham \ is the hypothesis that there is some deterministic
signal of interest, `$S$' in the observation \vym.

\begin{align}
  \label{eq:hypotheses}
  \begin{split}
  \hn &: \vy = \ \ \ \ \ \ \   \vw \\
  \ha &: \vy = S + \vw
  \end{split}
\end{align}

The minimum probability of error decision rule corresponds to the
picking the hypothesis with the maximum \emph{a~posteriori}
probability (MAP), Eq.\ref{eq:MAP},

\begin{equation}
  \hat{H}(\vy) = \underset{A\in \{\hn,\ha\}}{ \text{arg max} } Pr[H = A | \vy = y ]
  \label{eq:MAP}
\end{equation}

where $y$ is a realization of \vym.

For additive white Gaussian noise (AWGN), Eq.\ref{eq:GN},
\begin{equation}
  \vw \sim N(\mv,\sig I)
  \label{eq:GN}
\end{equation}

the MAP decision rule, Eq.\ref{eq:MAP} becomes Eq.\ref{eq:htestExpand}
\begin{equation}
  \ip{S}{\vy} \hypTest \frac{\ip{S}{S}}{2} + \ip{S}{\mv} +\sig \ln( \frac{P_{0}}{P_{1}})
  \label{eq:htestExpand}
\end{equation}
Here $\ip{a}{b}$ is shorthand for the inner product of $a$ with $b$
using the definition of inner product for whatever Hilbert space on
which $a$ and $b$ are defined. Letting $\gamma$ be the right hand side
of Eq. \ref{eq:htestExpand}, as in Eq.\ref{eq:gamma}, and moving it to
the left hand side of the inequality, gives
Eq. \ref{eq:htestSimplified}.

\begin{equation}
  \gamma \equiv \frac{\ip{S}{S}}{2} + \ip{S}{\mv} +\sig \ln( \frac{P_{0}}{P_{1}})
  \label{eq:gamma}
\end{equation}

\begin{align}
  \label{eq:htestSimplified}
  \begin{split}
    \ip{S}{\vy} - \gamma(S,\vw,P_{0},P_{1}) &\hypTest 0
  \end{split}
\end{align}

Eq.\ref{eq:htestSimplified} shows that a binary hypothesis test's
threshold $\gamma$ is dependent on the noise, \vwm.

\subsection{Binary Classification via a Single Neuron}
\begin{figure}[t]
\begin{center}
\fbox{\rule{0pt}{1.0in}
  \includegraphics[width=0.6\linewidth, height=1in]{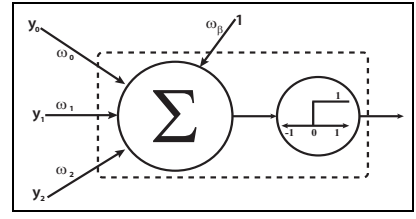}}
\end{center}
   \caption{A neuron consists of weighted inputs, a means of combination and an output non-linearity called an activation function.}
\label{fig:MPN}
\end{figure}

A single neuron with a Heavyside activation function,
Fig.\ref{fig:MPN}, implements a binary classifier
Eq.\ref{eq:neuronHeavySide}.

\begin{equation}
  \label{eq:neuronHeavySide}
\ip{\boldsymbol{\omega}}{\vy} + \omega_{\beta}  \overset{\hat{H}(\vy)=1}{\underset{\hat{H}(\vy)=0}{\gtreqless}} 0
\end{equation}

By pattern matching between Eq.\ref{eq:htestSimplified} and
Eq.\ref{eq:neuronHeavySide}, the weights $\boldsymbol{\omega}$ encode
the signal of interest `$S$' and the bias term, $\omega_{\beta}$,
corresponds to the threshold $\gamma$. Therefore, a neuron is a
machine which implements the MAP decision rule in the presence of
additive Gaussian noise. Since $\gamma$, Eq.\ref{eq:htestSimplified}, is
a function of the noise, \vwm, in order to maintain performance when
the noise conditions change, the neuron's bias must also change.

\subsection{Deviations from the model}
There are two major deviations from the theory. First, the signal
being looked for, $S$, encoded into the weights of a match filter also
changes with a change in the noise level. Second, the activation
function isn't a Heaviyside step function, and therefore, a neural
network isn't purely a binary machine.

The $S$ encoded into the neuron's weights vary with noise. This
follows from the fact that object features that are visible and can be
used for training when there is no noise, may be occluded when
training with noise. Therefore, the $S$s when training with noise and
without noise are not necessarily going to be the same. Since features
that are visible in the presence of noise will also be visible when
there is no noise, one can expect, and it has been observed, both here
an in \cite{Sietsma:1991:CAN:104793.104807}, that a network trained on
noisy inputs will stand up better to a lessening of noise level than a
clean trained network will to an increase in noise level.

Next, the weights, $\boldsymbol{\omega}$ are estimated from input data
used for training, \vytm. This means that the $S$ encoded by the
weights is not a known deterministic function, as had been assumed in
Sec.\ref{sec:BHT}. Instead, the $S$ encoded in the neuron's weights is
really $\hat{S}(\vyt)$, an estimate of the true $S$, so it is a
function of the input signal used during training $\vyt$ and thus a
function of the noise.

Also, in order to have continuous derivatives when doing back
propogation during training, the activation function is typically a
smooth differentiable function like a sigmoid or \emph{tanh}, not a
discontinous Heavyside step function. In this paper, we use
\emph{tanh}. This means that more information besides just present or
not present is being transmitted from the first convolutional layer to
the next. In this case, one might expect, and we show, that adjusting
the biases of the layers besides the input convolutional layer might
have an impact of performance. It would also be expected that the
inner layer's impact would decrease as the chosen activation function
becomes closer to the Heavyside function. Demonstrating this is for
future work.

\section{Proposed Method - Adjusting Biases}
Following directly from the theory in Sec.\ref{sec:BHT}, our method is
to simply adjust the biases of the input convolutional layer or layers
in occordinance with the noise in the input signal. This method
changes the threshold to the correct threshold for detection at that
level of noise. Comparing the output of our correlator to the correct
threshold dramatically improves the ability of the network to handle
changes in input noise. This kind of situation could occur in numerous
applications like automotive applications or simple street
sign/address recognintion during rain or fog. The noise could be
random occlusion caused by rain or snow or background noise caused by
bright sun or darkness.

Data about the amount of rain or snow falling in the area could be
acquired by direct measurement by onboard sensors and/or by
information received by the vehicle via some network. If a measured
noise level falls between two noise levels that were used for
training, interpolation can be used to create appropriate biases for
the measured noise level. If a measured noise level falls outside the
expected range, interpolation can once again be used, if it is safe to
do so.  Our method comprises both a training stage,
Sec.\ref{sec:Training}, and a runtime operation stage,
Sec.\ref{sec:operation}.

\subsection{Training}
\label{sec:Training}
\begin{figure}[t]
\begin{center}
\fbox{\rule{0pt}{2.0in}
   \includegraphics[width=0.8\linewidth, height=4.0in]{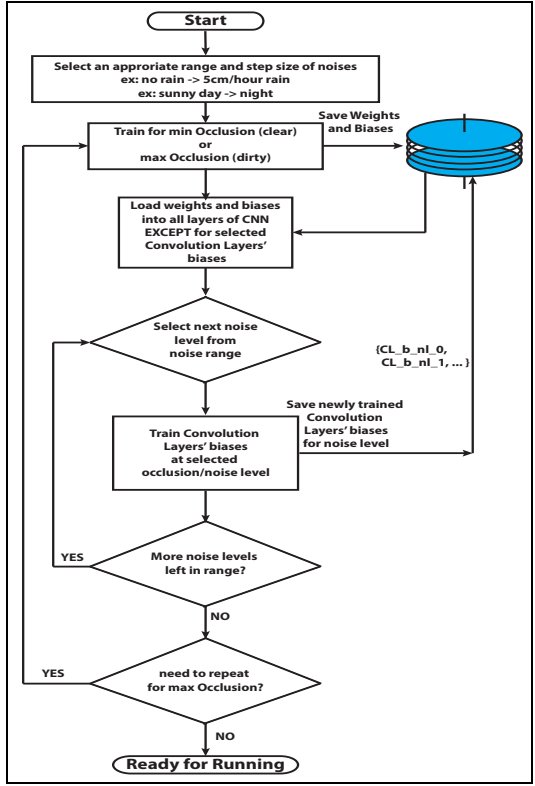}}
\end{center}
   \caption{A flow diagram of our training methodology.}
\label{fig:trainingFlow}
\end{figure}

Fig.\ref{fig:trainingFlow} shows the training flow.  In order to train
for noise, we need to select an appropriate range of expected noise
levels. The minimum expected noise level will be referred to as the
\emph{zero} or \emph{clean} noise level. In our tests, this is the
absence of any noise. The maximum expected noise level will be called
the \emph{max} or \emph{dirty} noise level. All network parameters are
trained for each of these two different noise levels. Then, the clean
parameters are loaded into the network and the network is trained for
each individual noise level. This time, only the biases on the input
convolution layer (or layers) are allowed to train and are saved for
each individual noise level. This is repeated for the \emph{max} noise
set. Each set of params, the \emph{zero} set and the \emph{max} set
will now have an associated set of biases for each noise level.

\subsubsection{Runtime Operation}
\label{sec:operation}
\begin{figure}[t]
\begin{center}
\fbox{\rule{0pt}{2.0in}
   \includegraphics[width=0.8\linewidth, height=4.0in]{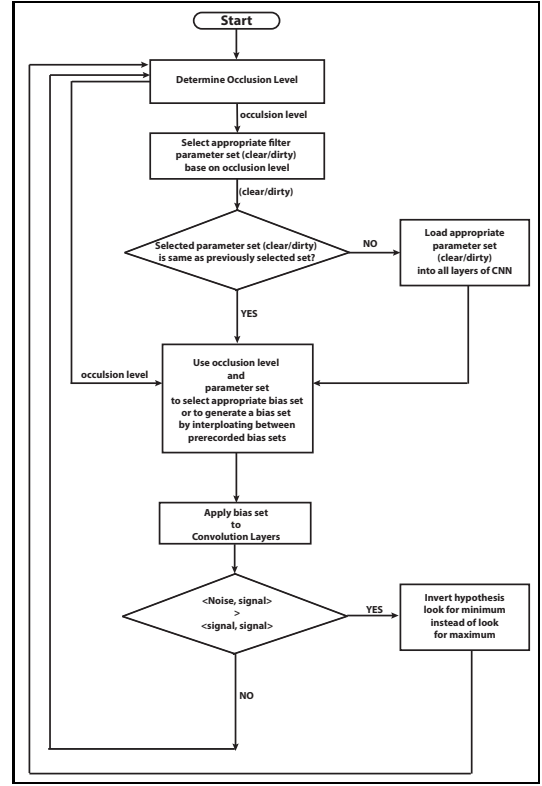}}
\end{center}
   \caption{A flow diagram of our running methodology.}
\label{fig:runtimeFlow}
\end{figure}

The runtime flow is shown in Fig.\ref{fig:runtimeFlow}. During
runtime, the ambient noise level is being continuously measured. This
is then used to first select the appropriate parameter set \emph{zero}
or \emph{max}. Then it is again used to select which set of biases
should be used for that noise level. If the measured noise does not
exactly match a bias, the nearest bias set can be used, or
interpolation can be used to quickly create a better fitting bias set.

Noise isn't always just zero mean AWGN. Sometimes it is non-zero mean
noise. Noise like this is more like camouflage. The ambient lighting
may make the background much lighter or darker. Thus making it more
difficult to distinguish the object from the background. This creates
three situations for our matched filters,

\begin{align}
  \label{eq:situations}
  \begin{split}
    \ip{S}{\bk} &\mathbf{=} \ip{S}{S} \\
    \ip{S}{\bk} &\mathbf{<} \ip{S}{S} \\
    \ip{S}{\bk} &\mathbf{>} \ip{S}{S} \\
  \end{split}
\end{align}

Where $\bk$ is the background of the image. When the background and
$S$ are similar in color and intensity, $\ip{S}{\bk} =
\ip{S}{S}$. This makes detection very difficult. When $\ip{S}{\bk} <
\ip{S}{S}$, the normal decision rule of Eq.\ref{eq:htestSimplified}
applies. However, when $\ip{S}{\bk} > \ip{S}{S}$, the hypothesis need
to be reversed as in Eq.\ref{eq:refHyp}.

\begin{align}
  \label{eq:refHyp}
  \begin{split}
    \ip{S}{\vy} - \gamma(S,\vw,P_{0},P_{1}) &\revhypTest 0
  \end{split}
\end{align}

%% file: results.tex
\section{Test Setup}
We used the LeNet model\cite{726791} for Theano
\cite{2016arXiv160502688full} that is avilable on the lisa-lab
DeepLearningTutorials \cite{DLT} from gitHub. This model is made up of
two convolutional layers, one hidden layer and a logistic regression
layer. The layers have 20 filters, 50 filters, 500 neurons and 10
neurons respectively.

\begin{figure}
\begin{center}
\fbox{\rule{0pt}{1.0in}
  \includegraphics[width=1.0\linewidth]{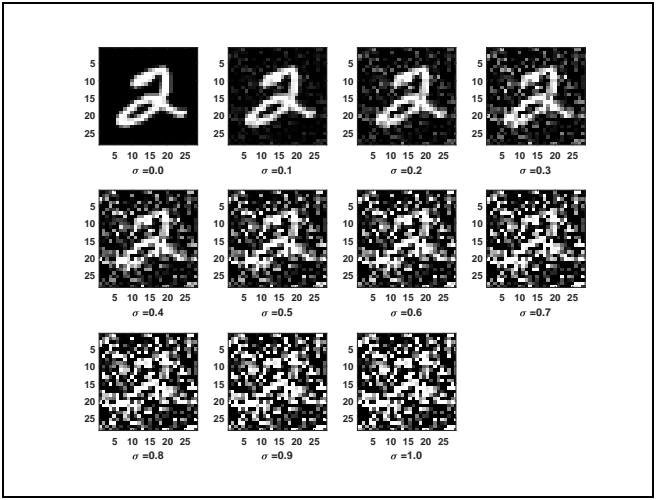}}
\end{center}
   \caption{The MNIST number 2 is corrupted with zero mean AWGN.}
\label{fig:NoisyNumbers}
\end{figure}

Each image in the MNIST dataset\cite{mnistlecun} is 28x28 pixels whose
intensity is in the range $[0,1]$. Each image was corrupted with AWGN
of a particular standard deviation. Multiple datasets were created,
one set for each noise with a specific standard deviation in the range
of $[0,1]$.  Modified pixels whose values exceeded either extreme were
clamped to that exterme value. Fig.\ref{fig:NoisyNumbers} shows a
representative set of images of the number 2 with increasing amounts
of AGWN.

We test four different kinds of networks: network trained at each
noise level, a mixed noise trained network, a zero noise trained
network, and a max noise trained network. All networks were trained on
a batch size 1000 with a learning rate of 0.05 and 1000 epochs
maximum.

\begin{figure}
\begin{center}
\fbox{\rule{0pt}{1.0in}
  \includegraphics[width=1.0\linewidth]{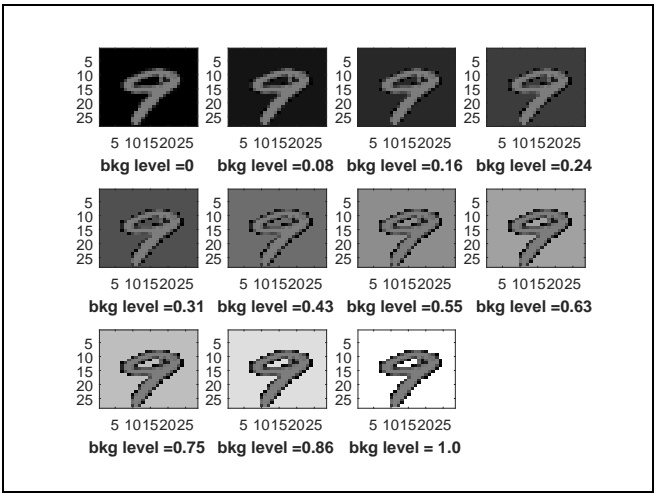}}
\end{center}
   \caption{The MNIST number 9 is corrupted with a uniform background
     noise of different intensities.}
\label{fig:backgroundNoise}
\end{figure}

Noise is not just restricted to zero mean Gaussian noise. Noise can
also come in the form of a scene being too bright or too
dark. Therefore, we also tested our system with MNIST numbers that
have a background that varies in intensity as seen in
Fig.\ref{fig:backgroundNoise}. These changing background level results
are seen in Sec.\ref{sec:cbl}.

Finally, we run tests were we preprocessed the noisy input images with
a denoising autoencoder (dA). Like the LeNet CNN, the dA used was
taken from the Theano tutorials. It was a single hidden layer
autoencoder with 500 hidden nodes. The dA was used as a preprocessor
to the aforementioned LeNet network. The dA, takes in a corrupted
image, and outputs a cleaned up version of that image. The cleaned up
image was then input into the LeNet network. The LeNet network was
trained using an unaltered MNIST database. The dA used at each noise
level was either a dA that had been trained for the \emph{zero} or
\emph{max} noise levels or for a mix of noise levels.
 
\section{Results}
\label{sec:results}
This section presents results for all comparisons. In all situations,
training for and testing on a specific noise level lead to the best
results. This method, however, may be too heavy on system resouces and
does not allow for the ability to modify the system when the measured
noise does not exactly match a trained noise level. Pruning may be
used, however, this can lead to more sensitivity to noise as was
observed in \cite{Sietsma:1991:CAN:104793.104807}.

\subsection{Impact of zero mean finite variance AWGN}
\begin{figure}
\begin{center}
  \fbox{\rule{0pt}{1.0in}
    \includegraphics[width=1.0\linewidth]{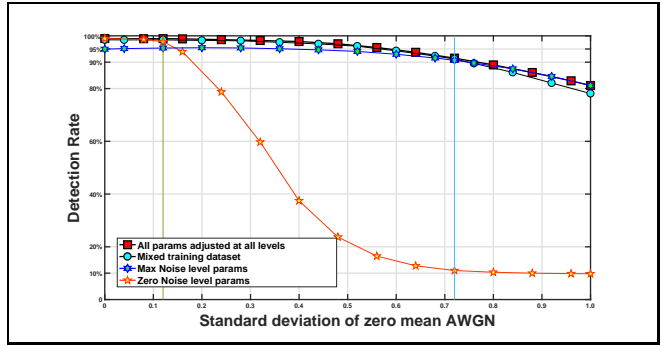}}
\end{center}
   \caption{The mixed trained network outperforms the zero noise and
     max noise networks in the range [0.12,0.72].}
\label{fig:fundamentalFig}
\end{figure}

Fig.\ref{fig:fundamentalFig} is a plot of standard deviation of AWGN
vs detection rate for the base methods: all parameters trained (and
therefore adjusted) at all noise levels, all parameters were trained
with a mixed noise dataset, a zero noise trained network and a max
noise trained network. The highest performing network at each noise
level is the network has been trained for that specific noise
level. This is denoted by the line with the square markers. The circle
line is for the network that was trained with a mix of data that was
drawn from the entire noise range. The hexagram line uses the
parameters from the maximum noise level trained network. The star line
uses the parameters for the zero noise level trained network.

As can be seen, training for each noise level maximizes performance at
that noise level, but this takes considerable storage space and
swapping out parameters on such a fine scale may be taxing on system
resources.

Training with the mixed noise level dataset creates the best
performing network over the widest range possible when using only a
single set of parameters. Since storage and memory bandwidth are at a
premium in a real time system, this is the most practical alternative
to our system. However, mixed noise training is not a performance
leader at either end of the noise spectrum. Using it would only be
optimal if there was a perpetual level of occlusion.

As seen in Fig.\ref{fig:fundamentalFig}, the zero noise trained
network out performs the mixed noise trained network until a standard
deviation of about 0.12. This is marked by a veritcal line. The max
noise network outperforms the mixed noise network in the standard
deviation noise range of [0.72,1.0]. This leaves a wide range, [0.12,
  0.72] in which the mixed noise trained network is the simplest and
most robust network. The 0.72 standard deviation is also accented by a
vertical line.

As expected, the maximum noise trained network holds up to a change in
its noise environment better than the zero noise trained network. As
stated earlier, this kind of behavior has been observed before
\cite{Sietsma:1991:CAN:104793.104807}.  This is reasonable because the
features learned when training with noise, if the noise is white
noise, should be there whether the noise is present or not. However,
the features learned when training without noise, may become corrupted
or totally obstructed in an image with noise.
 
\subsection{Adjusting input convolution layer biases}
\begin{figure}
  \begin{center}
\fbox{\rule{0pt}{1.0in}
   \includegraphics[width=1.0\linewidth]{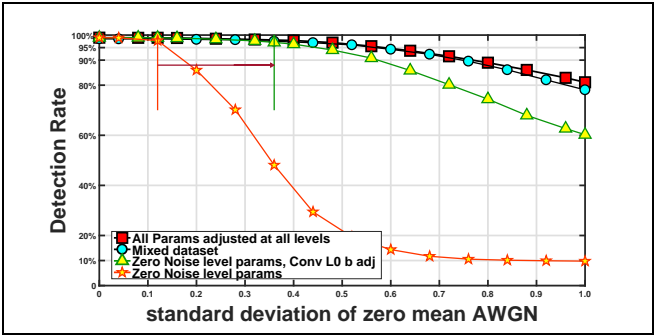}}
\end{center}
   \caption{Adjusting the biases of the input convolution layer
     extends the upper noise limit of the region where the zero noise
     level parameters out perform the mixed noise level parameters
     from 0.12 to 0.36. Here the zero noise parameters were used for
     testing.}
   \label{fig:cleanb0}
\end{figure}

Fig.\ref{fig:cleanb0} shows the impact of adjusting the input
convolution layer's biases when using the zero noise parameter
set. Adjusting the biases extends the range on which the zero noise
network outperforms the mixed network from 0.12 to around
0.36. Furthermore, the worst the zero noise network now performs
improves from around 10\% to around 60\%, approximately a 6x
improvement in performance.

\subsection{Adjusting multiple convolution layers biases}
\begin{figure}
  \begin{center}
\fbox{\rule{0pt}{1.0in}
   \includegraphics[width=1.0\linewidth]{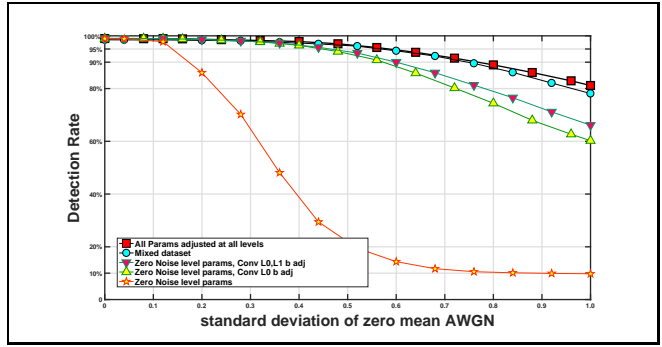}}
\end{center}
   \caption{Adjusting the biases of both convolution layers results in
     better performance.}
   \label{fig:cleanb01}
\end{figure}

Fig.\ref{fig:cleanb01} shows that adjusting both of the convolution
layer's biases leads to further improvements in detection
performance. This behavior is not expected when using Heavyside
activation functions. However, the activation function is $tanh(x)$,
so more information than just a binary result is being transmitted to
the next layer. This improvement is also understandable in that
adjusting the biases of the inner layers makes the network closer to
the best performing network that had been trained just for that layer.

\subsection{Adjusting Max noise network}
\begin{figure}
  \begin{center}
\fbox{\rule{0pt}{1.0in}
   \includegraphics[width=1.0\linewidth]{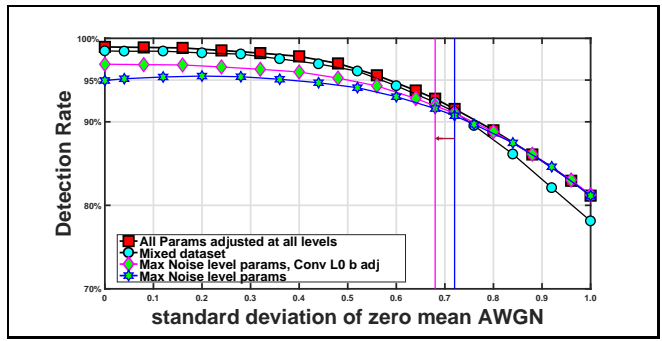}}
\end{center}
  \caption{Effect of adjusting noisy filters' biases.}
  \label{fig:dirtyb0}
\end{figure}

Fig.\ref{fig:dirtyb0} shows how the max noise network's performance
improves with bias adjustment. The region in which it beats the mixed
noise network goes from 0.72 to about 0.68, not as big of an
improvement as observed with the zero noise network.

\begin{figure}
  \begin{center}
\fbox{\rule{0pt}{1.0in}
   \includegraphics[width=1.0\linewidth]{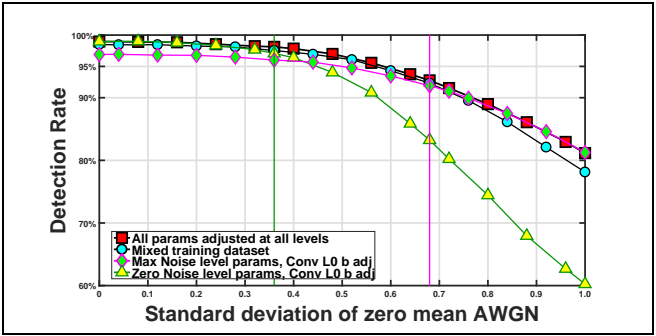}}
\end{center}
  \caption{Adjusting the biases of the zero and max noise networks
    significantly decreases the mixed noise trained network's optimal
    range from the range seen in Fig.\ref{fig:fundamentalFig}.}
  \label{fig:cleandirtyBiasAdjust}
\end{figure}

Fig.\ref{fig:cleandirtyBiasAdjust} recreates
Fig.\ref{fig:fundamentalFig}, but with the zero noise network and max
noise network's biases adjusted. This shrinks the area in which the
mixed training is better than adjusting biases from [0.12,0.72] to
[0.36, 0.68].

This results in a range of operation between 0.36 and 0.68 in which
the mixed noise network out performs the other two bias adjusted
networks.

\subsection{Changing background level}
\label{sec:cbl}
\begin{figure}
  \begin{center}
\fbox{\rule{0pt}{1.0in}
    \includegraphics[width=1.0\linewidth]{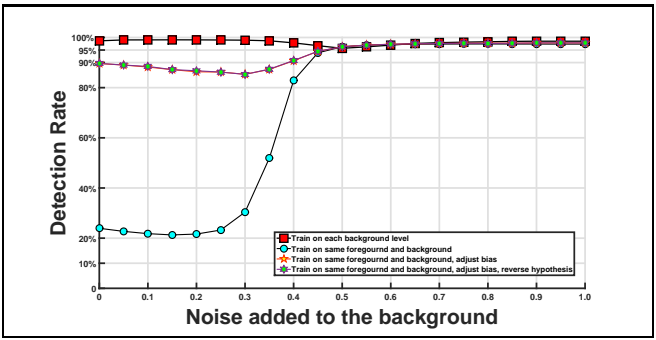}}
\end{center}
  \caption{Bias adjustment helps maintain detection rate. Noise level is the mean noise value and not the variance.}
  \label{fig:meanNoise}
\end{figure}

There are two situations to consider when changing the background
level. The first is that the noise is at the same level as the
interior. The first case results in Fig.\ref{fig:meanNoise}. Here,
rule reversal doesn't matter as the network has been trained on a mean
noise level of 0.5 and the body of each letter is also 0.5. This means
that, in order for detection to work, the network does not use the
character's body information. It only learns about the small
perturbations at the edge of the letter. Some of these perturbations
can be seen around the letters in Fig.\ref{fig:backgroundNoise}. By
inspection, it seems that these edge markers were somewhat more likely
to be dark than light, so this explains the drop in performance as the
background becomes more dark. Dark is indicated by a noise level of
0.0 and a clear background is a noise level of 1.0.

\begin{figure}
  \begin{center}
\fbox{\rule{0pt}{1.0in}
   \includegraphics[width=1.0\linewidth]{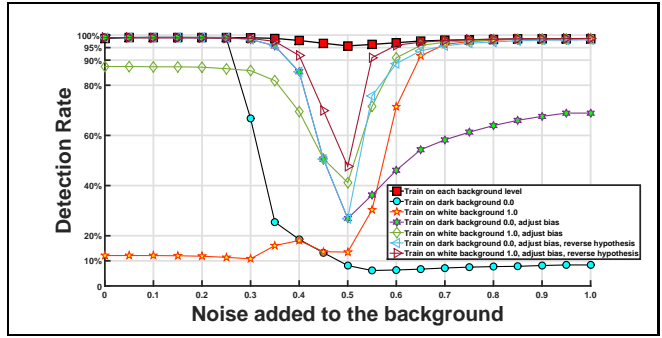}}
\end{center}
  \caption{The network has been trained at extremes of background
    levels. Both rule reversal and bias adjustment should be used
    in order to maintain performance.}
  \label{fig:noiseNoise}
\end{figure}

The second situation is when the network has been trained with one of
the two extremes of noise. The second situation results in
Fig.\ref{fig:noiseNoise}. In this case, if rule reversal is not
utilized, the performance of the networks drops dramatically when
approaching and after crossing the half way mark in the direction away
from the trained noise level. If rule reversal is used, then
performance drops as the ambient level approached the 0.5 mark and
then returns back up as the ambient level again diverges from the
level of the inner part of a character.

This suggests that training under ambient lighting conditions that are
about the same level as the object you are trying detect is the best
policy for minimizing the effects of changes in ambient lighting
conditions. This is to say that, one should try to camouflage the
object they are trying to detect when training. This again goes back
to the idea that features that are learnable in the presence of
camouflage are going to be there even when the camouflage is removed.

\subsection{Denoising AutoEncoder as preprocessor}
\begin{figure}
  \begin{center}
\fbox{\rule{0pt}{1.0in}
   \includegraphics[width=1.0\linewidth]{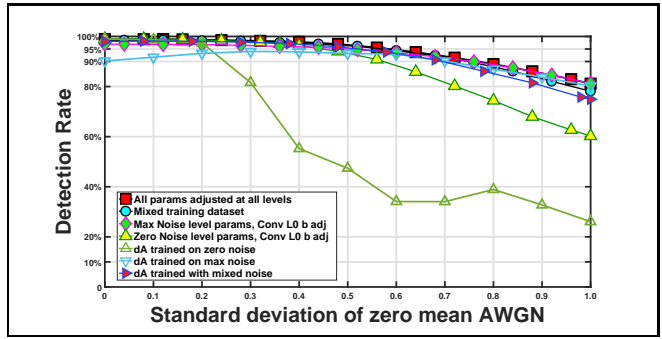}}
\end{center}
  \caption{Denoising autoencoder performs well over a large range when
    trained with a mixed dataset; however, it doesn't perform as well
    as our methoid in the extreme regions or the network itself just
    trained with mixed data.}
  \label{fig:dAcomparison}
\end{figure}

Denoising autoencoders can be used to remove noise from
images. However, dAs are very computationally intensive, so they may
not be well suited for an embedded environment. If the case of the dA,
once again, training without noise leads to the worst
performance. Training with noise gives performance slightly better
than our bias technique for the \emph{max} noise level in the low
noise regime. However, it does not match the performance of our
technique for the zero noise trained bias controlled network,
Fig.\ref{fig:dAcomparison}.

%% file: conclusions.tex
\section{Conclusions and Future Work}
Networks trained for individual noise levels seem to have the best
performance. Mixed noise trained networks perform well on average, but
not at the extremes, so such a network would be suboptimal when
conditions are good and possibly dangerously subobitmal when it is
needed the most in bad conditions. The noisy bias adjusted network
performs well when there is lots of noise, but the the clean trained
bias adjusted network peforms better when there is low levels of
noise. In short, there seems to be no optimal network for all noise
levels.

Also, because of their computational complexity, preprocessing a
signal with a denoising autoencoder does not look to be as effective
as one might initially hope.

Our technique of adjusting the biases of the convolutional layers
helps support detection performance in the presence of a change in
noise level from that with which the network was trained. It has been
shown to extend the effective range for both the zero noise parameters
and the max noise parameters.  Adjusting the biases is computationally
efficient and simple to implement and is this an attractive part of a
possible solution. How this technique can be used in a specific
situation will require further research.